\documentclass[11pt,a4paper]{article}
\usepackage[hyperref]{emnlp-ijcnlp-2019}
\usepackage{pslatex}
\usepackage{latexsym}
\usepackage{amsmath}
\usepackage{graphicx}
\usepackage{lipsum, color}
\usepackage{mathtools, cuted}
\usepackage{multirow}
\usepackage{array}
\usepackage{multicol}
\usepackage{url}
\usepackage{rotating}
\makeatletter
\g@addto@macro{\UrlBreaks}{\UrlOrds}
\makeatother
\usepackage{array}
\newcolumntype{L}[1]{>{\raggedright\let\newline\\\arraybackslash\hspace{0pt}}m{#1}}
\newcolumntype{C}[1]{>{\centering\let\newline\\\arraybackslash\hspace{0pt}}m{#1}}
\newcolumntype{R}[1]{>{\raggedleft\let\newline\\\arraybackslash\hspace{0pt}}m{#1}}
\usepackage{pifont}

\usepackage{amsmath,amsfonts,amssymb,amsthm}
\usepackage{mathtools}
\usepackage{commath}
\usepackage{xcolor}
\definecolor{tp1}{HTML}{66CC00}
\definecolor{tp2}{HTML}{009900}
\definecolor{tp3}{HTML}{999900}
\definecolor{tp4}{HTML}{FF0000}
\definecolor{tp5}{HTML}{FF6666}
\usepackage{subcaption}
\usepackage{caption}
\usepackage[inline]{enumitem}
\usepackage{caption, subcaption}

\usepackage{booktabs}
\newcommand{\tabitem}{~~\llap{\textbullet}~~}
\usepackage{xargs}     
\usepackage{color,soul}
\usepackage[colorinlistoftodos,prependcaption,textsize=scriptsize]{todonotes}

\aclfinalcopy 

\title{Movie Plot Analysis via Turning Point Identification}

\author{Pinelopi Papalampidi  \qquad
  Frank Keller \qquad
  Mirella Lapata \\
  Institute for Language, Cognition and Computation \\
    School of Informatics, University of Edinburgh \\
    10 Crichton Street, Edinburgh EH8 9AB \\
  {\tt p.papalampidi@sms.ed.ac.uk, \{keller,mlap\}@inf.ed.ac.uk }
    }

\date{}

\makeatletter
\newcommand{\thickhline}{%
    \noalign {\ifnum 0=`}\fi \hrule height 1pt
    \futurelet \reserved@a \@xhline
}
\makeatother

\begin{document}

\maketitle

\begin{abstract}
 
  According to screenwriting theory, turning points (e.g.,~change of
  plans, major setback, climax) are crucial narrative moments within a
  screenplay: they define the plot structure, determine its
  progression and thematic units (e.g.,~setup, complications,
  aftermath).  We propose the task of turning point identification in
  movies as a means of analyzing their narrative structure. We argue
  that turning points and the segmentation they provide can facilitate
  processing long, complex narratives, such as screenplays, for
  summarization and question answering. We introduce a dataset
  consisting of screenplays and plot synopses annotated with turning
  points and present an end-to-end neural network model that
  identifies turning points in plot synopses and projects them onto
  scenes in screenplays. Our model outperforms strong baselines based
  on state-of-the-art sentence representations and the expected
  position of turning points.
\end{abstract}

\section{Introduction}

Computational literary analysis works at the intersection of natural
language processing and literary studies, aiming to evaluate various
theories of storytelling (e.g.,~by examining a collection of works
within a single genre, by an author, or topic) and to develop tools
which aid in searching, visualizing, or summarizing literary content.

Within natural language processing, computational literary analysis
has mostly targeted works of fiction such as novels, plays, and
screenplays. Examples include analyzing characters, their
relationships, and emotional trajectories
\cite{chaturvedi2017unsupervised,iyyer2016feuding,Elsner:ea:2012},
identifying enemies and allies \cite{Nalisnick:Baird:2013}, villains
or heroes \cite{bamman2014bayesian,bamman2013learning}, measuring the
memorability of quotes \cite{Danescu-Niculescu-Mizil:ea:2012},
characterizing gender representation in dialogue
\cite{agarwal-EtAl:2015:NAACL-HLT,Ramakrishna:ea:2005,sap-etal-2017-connotation},
identifying perpetrators in crime series \cite{frermann2018whodunnit},
summarizing screenplays \cite{gorinski2018s}, and answering questions
about long and complex narratives \cite{kovcisky2018narrativeqa}.

\begin{table}[t]
\small
\centering
\begin{tabular}{@{}p{0.36\columnwidth}@{~~}p{0.62\columnwidth}@{}}
\thickhline
\multicolumn{1}{c}{{Turning Point}} & \multicolumn{1}{c}{{Description}} \\ \thickhline
{\color{tp1}{1. Opportunity}} & Introductory event that occurs after the presentation of the setting and the background of the main characters. \\ \thickhline

{\color{tp2}{2. Change of Plans}} & Event where the main goal of the story is defined. From this point on, the action begins to increase. \\\thickhline
{\color{tp3}{3. Point of No Return}} & Event that pushes the main character(s) to fully commit to their goal. \\\thickhline
{\color{tp4}{4. Major Setback}} & Event where everything falls apart (temporarily or permanently). \\\thickhline
{\color{tp5}{5. Climax}} & Final event of the main story, moment of resolution and the ``biggest spoiler''. \\ \thickhline
\end{tabular}
\caption{Turning points and their definitions.}
\label{tab:tp_description}
\end{table}

In this paper we are interested in the automatic analysis of narrative
structure in screenplays. Narrative structure, also referred to as a
storyline or plotline, describes the framework of how one tells a
story and has its origins to Aristotle who defined the basic
triangle-shaped plot structure representing the beginning (protasis),
middle (epitasis), and end (catastrophe) of a story
\cite{pavis1998dictionary}.  The German novelist and playwright Gustav
Freytag modified Aristotle's structure by transforming the triangle
into a pyramid \cite{freytag1896freytag}. In his scheme, there are
five acts (\textit{introduction, rising movement, climax, return}, and
\textit{catastrophe}).
Several variations of Freytag's pyramid are used today in film
analysis and screenwriting \cite{cutting2016narrative}.

In this work, we adopt a variant commonly employed by screenwriters as
a practical guide for producing successful screenplays
\cite{Hauge:2017}. According to this scheme, there are six stages
(acts) in a film, namely \textit{the setup}, \textit{the new
  situation}, \textit{progress}, \textit{complications and higher
  stakes}, \textit{the final push}, and \textit{the aftermath},
separated by five \emph{turning points} (TPs).  TPs are narrative
moments from which the plot goes in a different direction
\cite{thompson1999storytelling}, and by definition they occur at the
junctions of acts. Aside from changing narrative direction, TPs define
the movie's structure, tighten the pace, and prevent the narrative
from drifting. The five TPs and their definitions are given in
Table~\ref{tab:tp_description}.

 \begin{figure}[t]
   \centering
    \includegraphics[width=\columnwidth]{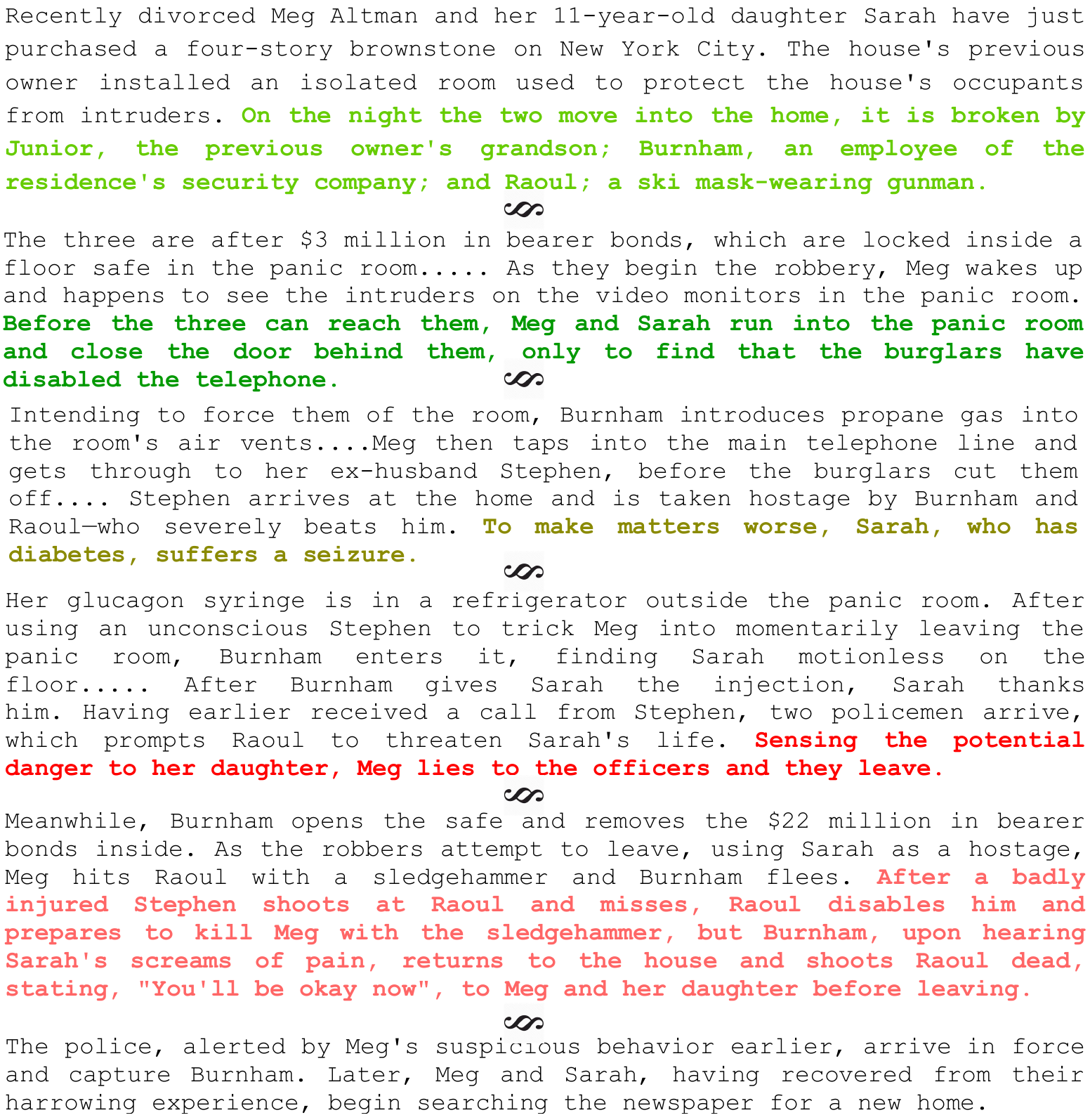}
    \caption{Example of turning point annotations ({\color{tp1}{TP1}},
      {\color{tp2}{TP2}}, {\color{tp3}{TP3}}, {\color{tp4}{TP4}},
      {\color{tp5}{TP5}}, respectively) for the synopsis of the movie
      ``Panic Room''.}
    \label{fig:tps_example}
\end{figure}

We propose the task of turning point identification in movies as a
means of analyzing their narrative structure. TP identification
provides a sequence of key events in the story and segments the
screenplay into thematic units.  Common approaches to summarization
and QA of long or multiple documents
\cite{chen2017reading,yang2018hotpotqa,kratzwald2018adaptive,elgohary2018dataset}
include a retrieval system as the first step, which selects a subset
of relevant passages for further processing. However,
\newcite{kovcisky2018narrativeqa} demonstrate that these approaches do
not perform equally well for extended narratives, since individual
passages are very similar and the same entities are referred to
throughout the story. We argue that this challenge can be addressed by
TP identification, which finds the most important events and segments
the narrative into thematic units. Downstream processing for
summarization or question answering can then focus on those segments
that are relevant to the task.

Problematically for modeling purposes, TPs are latent in screenplays,
there are no scriptwriting conventions (like character cues or scene
headings) to denote where TPs occur, and their exact manifestation
varies across movies (depending on genre and length), although there
are some rules of thumb indicating where to expect a TP (e.g.,~the
Opportunity occurs after the first 10\% of a screenplay, Change of
Plans is approximately 25\% in). To enable automatic TP
identification, we develop a new dataset which consists of
screenplays, plot synopses, and turning point annotations. To save
annotation time and render the labeling task feasible, we collect TP
annotations at the plot synopsis level (synopses are a few paragraphs
long compared to screenplays which are on average 120~pages long). An
example is given in Figure~\ref{fig:tps_example}. We then project the
TP annotations via distant supervision onto screenplays and propose an
end-to-end neural network model which identifies TPs in full length
screenplays.

Our contributions can be summarized as follows: (a)~we introduce TP
identification as a new task for the computational analysis of screen
plays that can benefit applications such as QA and summarization;
(b)~we create and make publicly available the
\textbf{T}u\textbf{R}n\textbf{I}ng \textbf{PO}int \textbf{D}ataset
(TRIPOD) \footnote{\url{https://github.com/ppapalampidi/TRIPOD}}
which contains 99 movies (3,329 synopsis sentences and 13,403
screenplay scenes) annotated with TPs; and (c)~we present an
end-to-end neural network model that identifies turning points in plot
synopses and projects them onto scenes in screenplays, outperforming
strong baselines based on state-of-the-art sentence representations
and the expected position of TPs.

\section{Related Work} 
\label{sec:related_work}

Recent years have seen increased interest in the automatic analysis of
long and complex narratives. Specifically, Machine Reading
Comprehension (MRC) and Question Answering (QA) tasks are
transitioning from investigating single short and clean articles or
queries \cite{rajpurkar2016squad,nguyen2016ms,trischler2016newsqa} to
large scale datasets that consist of complex stories
\cite{tapaswi2016movieqa,frermann2018whodunnit,kovcisky2018narrativeqa,joshi2017triviaqa}
or require reasoning across multiple documents
\cite{welbl2018constructing,wang2018multi,dua2019drop,yang2018hotpotqa}.
\newcite{tapaswi2016movieqa} introduce a multi-modal dataset
consisting of questions over 140 movies, while
\newcite{frermann2018whodunnit} attempt to answer a single question,
namely who is the perpetrator in 39 episodes of the well-known crime
series CSI, again based on multi-modal information. Finally, \newcite{kovcisky2018narrativeqa} recently introduced a dataset consisting of question-answer pairs over 1,572 movie screenplays and books.

Previous approaches have focused on fine-grained story analysis, such
as inducing character types \cite{bamman2013learning,
  bamman2014bayesian} or understanding relationships between
characters \cite{iyyer2016feuding,chaturvedi2017unsupervised}.
Various approaches have also attempted to analyze the goal and
structure of narratives. \newcite{black1979evaluation} evaluate the
functionality of story grammars in story understanding,
\newcite{Elson:McKeown:2009} develop a platform for representing and
reasoning over narratives, and \newcite{chambers2009unsupervised}
learn fine-grained chains of events.

In the context of movie summarization, \newcite{gorinski2018s}
automatically generate an overview of the movie's genre, mood, and
artistic style based on screenplay
analysis. \newcite{gorinski2015movie} summarize full length
screenplays by extracting an optimal chain of scenes via a graph-based
approach centered around the characters of the movie. A similar
approach has also been adopted by \newcite{vicol2018moviegraphs}, who
introduce the MovieGraphs dataset consisting of 51 movies and describe video clips with character-centered graphs. Other work creates animated
story-boards using the action descriptions of screenplays
\cite{Ye:Baldwin:2008}, extracts social networks from screenplays
\cite{agarwal-EtAl:2014:CLFL}, or creates \emph{xkcd} movie narrative
charts \cite{agarwal:ea:2014}.

Our work also aims to analyze the narrative structure of movies, but
we adopt a high-level approach. We advocate TP identification as a
precursor to more fine-grained analysis that unveils character
attributes and their relationships. Our approach identifies key
narrative events and segments the screenplay accordingly; we argue
that this type of preprocessing is useful for applications which might
perform question answering and summarization over
screenplays. Although our experiments focus solely on the textual
modality, turning point analysis is also relevant for multimodal tasks
such as trailer generation and video
summarization.

\section{The TRIPOD Dataset}
\label{sec:dataset}

The TRIPOD dataset contains 99 screenplays, accompanied with cast
information (according to IMDb), and Wikipedia plot synopses annotated
with turning points.  The movies were selected from the Scriptbase
dataset \cite{gorinski2015movie} based on the following criteria:
\begin{enumerate*}[label={(\alph*)}]
\item maintaining a variation across different movie genres
  (e.g.,~action, romance, comedy, drama) and narrative types
  (e.g.,~flashbacks, time shifts); and 
\item including screenplays that are faithful to the released movies
  and their synopses as much as possible.
\end{enumerate*}
In Table~\ref{tab:original_dataset}, we present various statistics of
the dataset.

Our motivation for obtaining TP annotations at the synopsis level
(coarse-grained), instead of at the screenplay level (fine-grained)
was twofold. Firstly, on account of being relatively short, synopses are
easier to annotate than full-length screenplays, allowing us to scale
the dataset in the future. Secondly, we would expect synopsis-level
annotations to be more reliable and the degree of inter-annotator
agreement higher; asking annotators to identify precisely where a
turning point occurs might seem like looking for a needle in a
haystack.
An example of a synopsis with TP annotations is shown in
Figure~\ref{fig:tps_example} for the movie ``Panic Room''.  Each TP is
colored differently, and both the chain of key events (colored text) and
resulting segmentation (\begin{sideways}{\S}\end{sideways}) are illustrated.

\begin{table}[t]
\small
\centering
\begin{tabular}{lrr}
\thickhline
 & {\begin{tabular}[c]{@{}c@{}}Train \end{tabular}} & {\begin{tabular}[c]{@{}c@{}}Test \end{tabular}} \\ \thickhline 
movies & 84 & 15 \\
turning points & 420 & 75 \\
synopsis sentences & 2,821 & 508 \\
screenplay scenes & 11,320 & 2,083 \\
\begin{tabular}[c]{@{}l@{}}synopsis vocabulary \end{tabular} & 7.9k & 2.8k \\
\begin{tabular}[c]{@{}l@{}}screenplay vocabulary \end{tabular} & 37.8k & 16.8k \\ \thickhline 
\multicolumn{3}{c}{\textit{per  synopsis}} \\ \thickhline
tokens & 729.8 (165.5) & 698.4 (187.4) \\ 
sentences & 35.4 (8.4) & 33.9 (9.9) \\
sentence tokens & 20.6 (9.5) & 20.6 (9.3) \\ \thickhline 
\multicolumn{3}{c}{\textit{per screenplay}} \\ \thickhline
tokens & \begin{tabular}[c]{@{}l@{}} 23.0k (6.6)\end{tabular} & \begin{tabular}[c]{@{}l@{}} 20.9k (4.5) \end{tabular} \\
sentences & 3.0k (0.9) & 2.8k (0.6) \\
scenes & 133.0 (61.1) & 138.9 (50.7) \\ \thickhline 
\multicolumn{3}{c}{\textit{per scene}} \\ \thickhline
tokens & 173.0 (235.0) & 150.5 (198.3) \\ 
sentences & 22.2 (31.5) & 19.9 (26.9) \\ 
sentence tokens & 7.8 (6.0) & 7.6 (6.4) \\ \thickhline
\end{tabular}
\caption{Statistics of the TRIPOD dataset; all means are shown with
  standard deviation in brackets.}
\label{tab:original_dataset}
\end{table}

In an initial pilot study, the three authors acted as annotators for
identifying TPs in movie synopses. They selected exactly one sentence
per TP, under the assumption that all TPs are present. Based on the
pilot, annotation instructions were devised and an annotation tool was
created which allows to label synopses with TPs
sentence-by-sentence. After piloting the annotation scheme on 30
movies, two new annotators were trained using our instructions and in
a second study, they doubly annotated five movies. The remaining movies in
the dataset were then single annotated by the new annotators.

We computed inter-annotator agreement using two different metrics:
\begin{enumerate*}[label={(\alph*)}]
\item total agreement (TA), i.e.,~the percentage of TPs that two
  annotators agree upon by selecting the exact same sentence; and
\item annotation distance, i.e., the distance $d[p_i, tp_i]$ between
  two annotations for a given TP, normalized by synopsis length:
\end{enumerate*}
\begin{align}
    d[p_i, tp_i] = \frac{1}{N}|p_i - tp_{i}| \label{eq:distance_1} 
\end{align}
where $N$ is the number of synopsis sentences and $tp_{i}$ and $p_i$
are the indices of the sentences labeled with TP~$i$ by two
annotators. The mean annotation distance $D$ is then computed by
averaging distances $d[p_i, tp_i]$ across all annotated TPs.

The TA between the two annotators in our second study was~64.00\% and
the mean annotation distance was 4.30\% (StDev~3.43\%).
The annotation distance per TP is presented in
Table~\ref{tab:plot_results_per_tp} (last line), where it is compared
with the automatic TP identification results (to be explained later).

\begin{figure*}[t]
        \tiny
        \centering
        \begin{subfigure}[b]{0.57\textwidth}
            \tiny
            \centering
            \includegraphics[width=\textwidth,page=1]{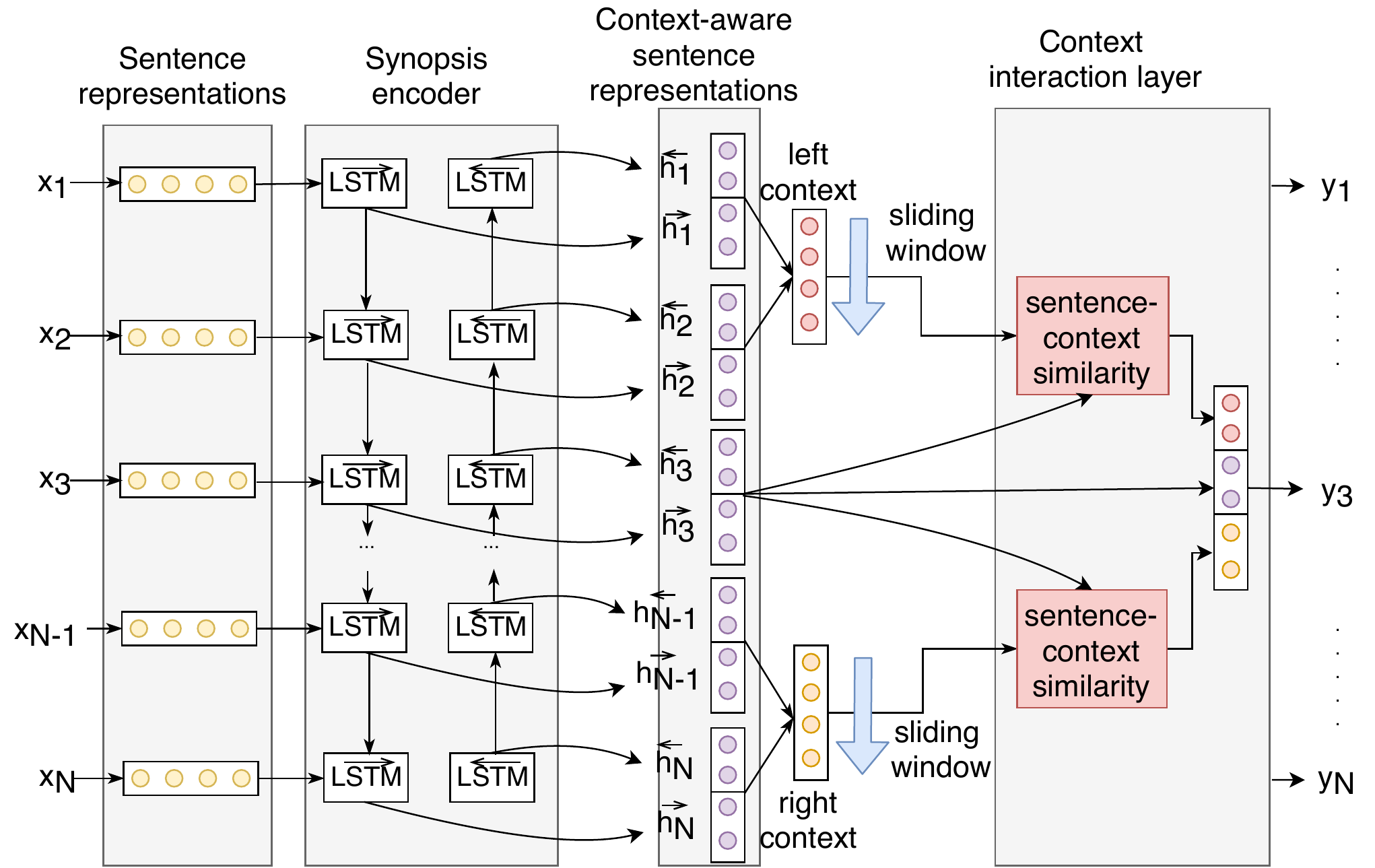}
            \caption[]%
            {{\small Topic-Aware Model (TAM)}}    
            \label{fig:context}
        \end{subfigure}
        \begin{subfigure}[b]{0.38\textwidth}  
             \tiny
            \centering 
            \includegraphics[width=\textwidth]{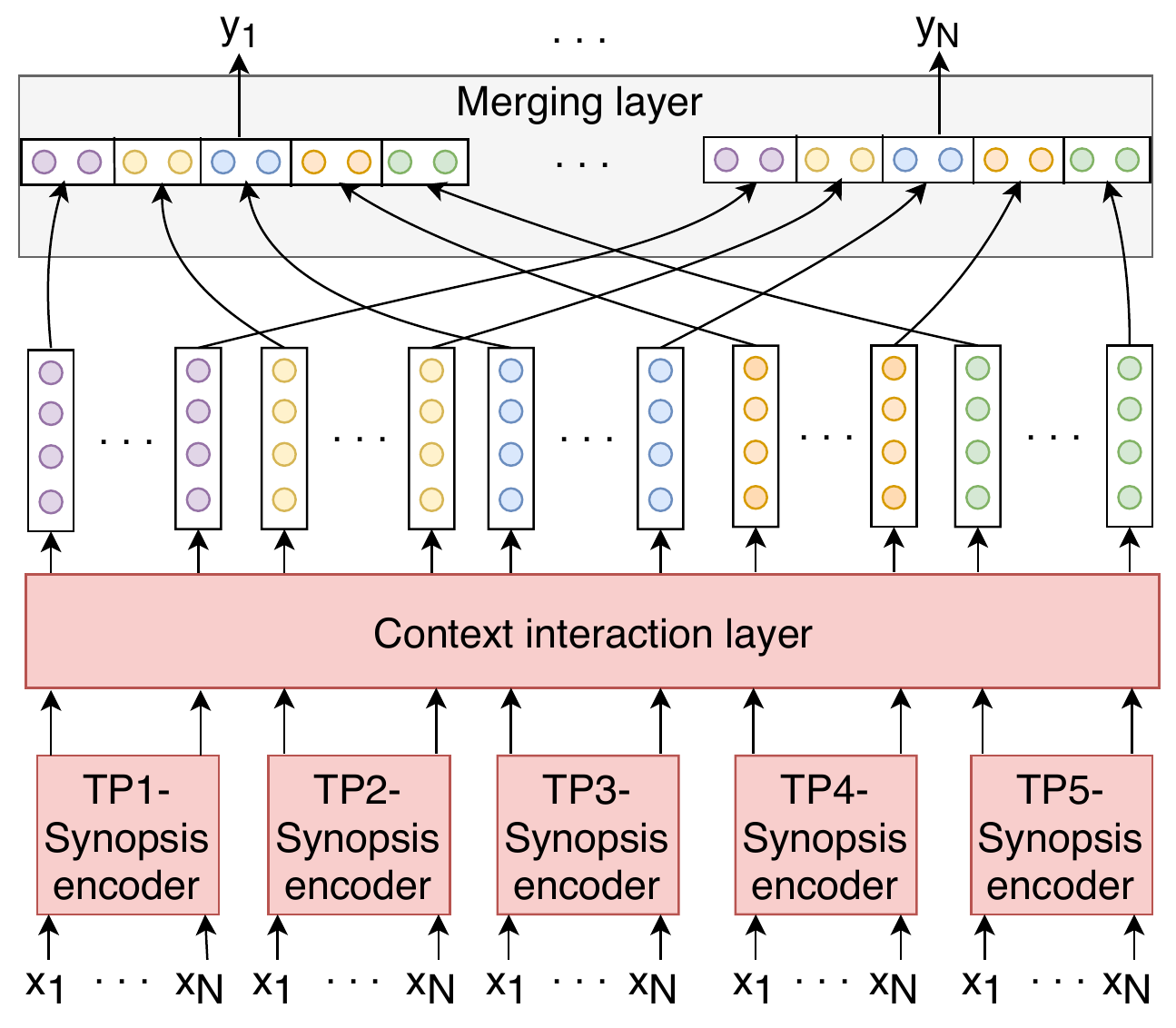}
            \caption[]%
            {{\small Multi-view TAM}}    
            \label{fig:tp-specific}
        \end{subfigure}
        \vspace{1.5em}
        \caption[ ]
        {\small Model overview for TP identification in synopses. On
          the left, sentence representations $x_i$ are contextualized
          via a synopsis encoder (BiLSTM layer) and after interacting
          with the left and right  windows in the context
          interaction layer, the final sentence representation $y_i$
          is computed. On the right, five different synopsis encoders
          are utilized, one per TP, and these different views of a
          synopsis sentence $x_i$ are combined in the merging layer.}
        \label{fig:model_plots}
\end{figure*}

We also asked our annotators to annotate the screenplays (rather than
synopses) for a subset of 15 movies. This subset serves as our
goldstandard test set. Annotators were given synopses annotated with
TPs and were instructed to indicate for each TP which scenes in the
screenplay correspond to it. Six of the 15 movies were doubly
annotated, so that we could measure agreement. Since annotators were
allowed to choose a variable number of scenes for each TP, this
changes slightly our agreement metrics.

Total Agreement (TA) now is the percentage of TP scenes the annotators
agree on:
\begin{align}
    TA = \frac{1}{T \cdot L}\sum_{i=1}^{T \cdot L}{\frac{|S_i \cap
      G_i|}{|S_i \cup G_i|}} \label{eq:total_agreement}
\end{align}
where $T$, $L$ are the TPs identified per annotator in a screenplay,
and~$S_i$ and~$G_i$ are the indices of the scenes selected for TP~$i$
by the two annotators. Partial Agreement (PA) is the percentage of TPs
where there is an overlap of at least one scene:
\begin{align}
    PA = \frac{1}{T \cdot L}\sum_{i=1}^{T \cdot L}{\left[ S_i \cap G_i \neq
      \emptyset \right]} \label{eq:partial_agreement}
\end{align}
And annotation distance~$D$ becomes the mean of the distances\footnote{We
  compute the minimum distance between the two sets of scenes, since
  non-sequential scenes may be included in the same set. Hence,
  considering the center of the sets is not always representative of
  the TP scenes.} $d[S_i, G_i]$ between two annotators normalized
by~$M$, the length of the screenplay:
\begin{gather}
    d[S_i, G_i] = \frac{1}{M} \min_{(s \in S_i, g \in G_i)} |s - g|
       \label{eq:distance_script_1} 
\end{gather}
The TA and PA between the two annotators were 35.48\% and 56.67\%,
respectively. The mean annotation distance was 1.48\%
(StDev~2.93\%). The TA shows that the annotators rarely indicate the
same scenes, even if they are asked to annotate an event in the
screenplay that is described by a specific synopsis sentence. However,
they identify scenes which are in close proximity in the screenplay,
as PA and annotation distance reveal. This analysis validates our
assumption that annotating the synopses first limits the degree of
overall disagreement.

\section{Turning Point Prediction Models} 
\label{sec:methodology}

In this work, we aim to detect text segments which act as TPs. We
first identify which \emph{sentences} in plot synopses are TPs
(Section~\ref{methodology:subtask1}); next, we identify which
\emph{scenes} in screenplays act as TPs via projection of
\emph{goldstandard} TP labels (Section~\ref{methodology:subtask2});
finally, we build an end-to-end system which identifies TPs in
screenplays based on \emph{predicted} TP synopsis labels
(Section~\ref{methodology:subtask3}).

All models we propose in this paper have the same basic structure; they
take text segments~$i$ (sentences or scenes) as input and predict
whether these act as TPs or not. Since the sequence, number, and
labels of TPs are fixed (see Table~\ref{tab:tp_description}), we treat
TP identification as a binary classification problem (where~1
indicates that the text is a TP and 0~otherwise).  Each segment is
encoded into a multi-dimensional feature space~$x_i$ which serves as
input to a fully-connected layer with a single neuron representing the
probability that~$i$ acts as a TP. In the following, we describe
several models which vary in the way input segments are encoded.

\subsection{Identifying Turning Points in  Synopses}
\label{methodology:subtask1}

\paragraph{Context-Aware Model (CAM)}
\label{single_plot_encoder}

A simple baseline model would compute the semantic representation of
each sentence in the synopsis using a pre-trained sentence
encoder. However, classifying segments in isolation without
considering the context in which they appear, might yield inferior
semantic representations. We therefore obtain richer representations
for sentences by modeling their surrounding context. We encode the
synopsis with a Bidirectional Long Short-Term Memory (BiLSTM;
\citealt{hochreiter1997}) network; and obtain contextualized
representation~$cp_i$ for sentence~$x_i$ by concatenating the hidden
layers of the forward~$\overrightarrow{h_i}$ and
backward~$\overleftarrow{h_i}$ LSTM, respectively:
$ cp_i = h_i = [\overrightarrow{h_i} ; \overleftarrow{h_i}]$ 
(for a more detailed description, see the Appendix). Representation
$cp_i$ is the input feature vector for our binary classifier. The
model is illustrated in Figure~\ref{fig:context}.

\paragraph{Topic-Aware Model (TAM)}
\label{sec:context_layer}

TPs by definition act as boundaries between different thematic units
in a movie. Furthermore, long documents are usually comprised of
topically coherent text segments, each of which contains a number of
text passages such as sentences or paragraphs
\cite{Salton:ea:1996}. Inspired by text segmentation approaches
\cite{hearst1997texttiling} which measure the semantic similarity
between sequential context windows in order to determine topic
boundaries, we enhance our representations with a \emph{context
  interaction} layer. The objective of this layer is to measure the
similarity of the current sentence with its preceding and following
context, thereby encoding whether it functions as a boundary between
thematic sections. The enriched model with the context interaction
layer is illustrated in Figure~\ref{fig:context}.

After calculating contextualized sentence representations $cp_i$, we
compute the representation of the left $lc_i$ and right $rc_i$
contexts of sentence~$i$ (see Figure~\ref{fig:context}, right-hand
side). We select windows of fixed length $l$ and calculate~$lc_i$ and
$rc_i$ by averaging the sentence representations within each window.
Next, we compute the semantic similarity of the current sentence with
each context representation. Specifically, we consider the
element-wise product $b_i$, cosine similarity $c_i$ and pairwise
distance $u_i$ as similarity metrics:
\begin{gather}
    b_i = cp_i \odot lc_i \quad\quad
    c_i = \frac{cp_i \cdot lc_i}{\norm{cp_i}\norm{lc_i}} \label{interaction2} \\
    u_i = \dfrac{cp_i \cdot lc_i}{\max(\Vert cp_i \Vert _2 \cdot \Vert lc_i \Vert _2)} \label{interaction3} 
\end{gather}
The interaction representation of sentence $cp_i$ with its left
context is the concatenation of $cp_i$, $fl_i$, and the above
similarity values (i.e., $b_i, c_i, u_i$):
\begin{gather} \label{interaction4}
fl_i = [cp_i ; lc_i ; b_i ; c_i ; u_i]
\end{gather}
The interaction representation $fr_i$ for the right context $rc_i$ is
computed analogously.  We obtain the final representation of
sentence~$i$ via concatenating $fl_i$ and $fr_i$:
$y_i = [fl_i ; fr_i ; cp_i]$.

\paragraph{TP-Specific Information}

Another variation of our model is to use TP-specific encoders instead
of a single one (see Figure~\ref{fig:tp-specific}). In this case, we
employ five different encoders for calculating five different
representations for the current synopsis sentence $x_i$, each one with
respect to a specific TP. These representations can be considered
multiple views of the same sentence. We calculate the interaction of
each view with the left and right context window, as previously, via
the context interaction layer. Finally we compute the sentence
representation $y_i$ by concatenating its individual context-enriched
TP representations.

\paragraph{Entity-Specific Information} \label{sec:es_encoder}

We also enrich our model with information about entities. We first
apply co-reference resolution to the plot synopses using the Stanford
CoreNLP toolkit \cite{manning2014stanford} and substitute mentions of
named entities whenever these are included in the IMDb cast list. We
then obtain entity-specific sentence representations as follows.  Our
encoder uses a word embedding layer initialized with pre-trained
entity embeddings and a BiLSTM for contextualizing word
representations.  We add an attention mechanism on top of the LSTM,
which assigns a weight to each word representation. We compute the
entity-specific representation $e_i$ for synopsis sentence~$i$ as the
weighted sum of its word representations (for more details, see the
Appendix). Finally, entity enriched sentence representations $x_i'$
are obtained by concatenating generic vectors $x_i$ with
entity-specific ones~$e_i$: $x_i' = [x_i ; e_i]$.

\begin{figure}[t]%
    \centering
    \includegraphics[width=0.5\textwidth]{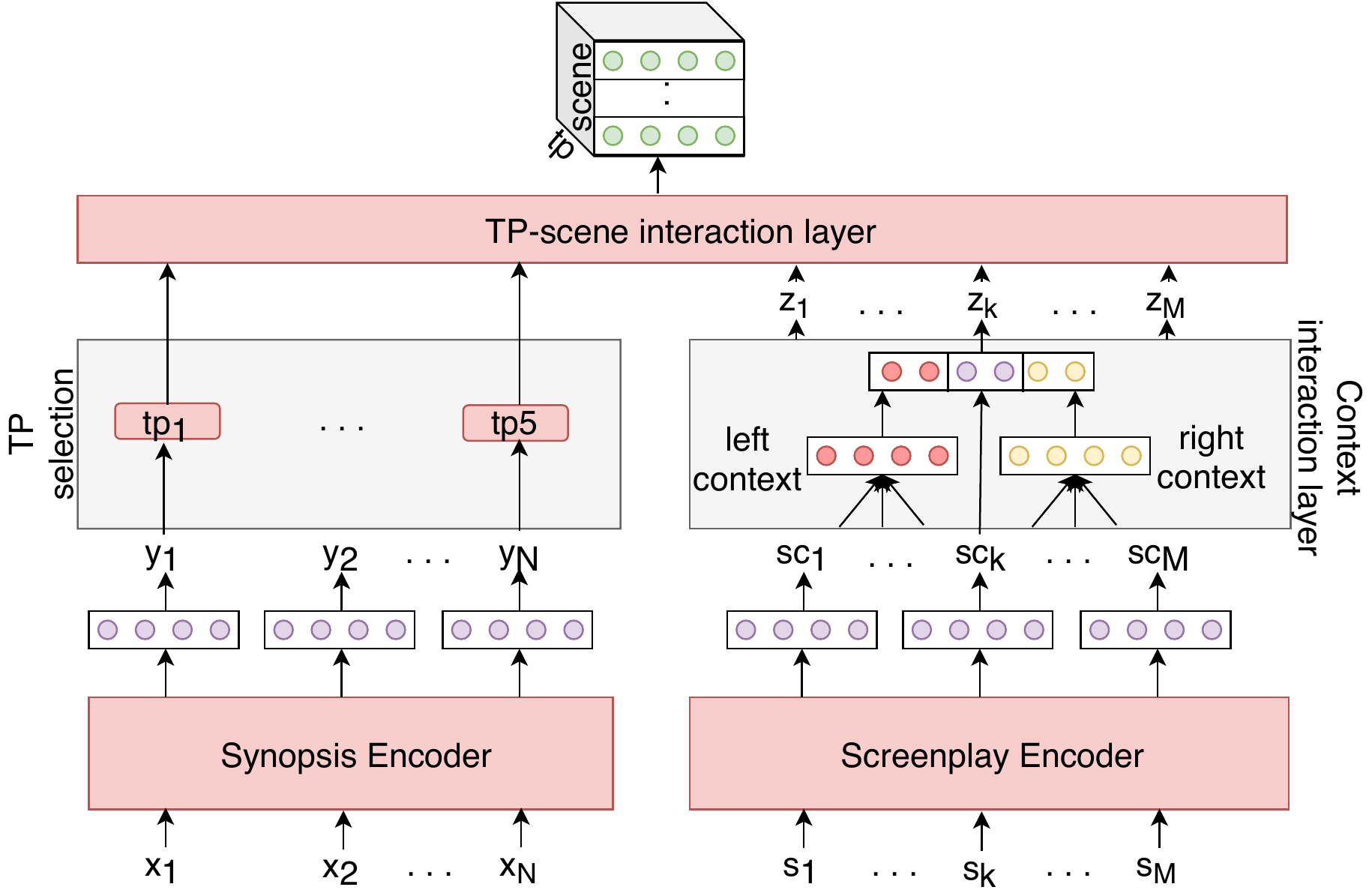}
    \caption{TAM overview for TP identification in screenplays.  The
      synopsis and screenplay encoders contextualize synopsis
      sentences~$x_i$ and screenplay scenes~$s_i$, respectively. TPs
      are selected from contextualized synopsis sentences $y_i$ and a
      richer representation $sc_i$ is computed for~$s_i$ via the
      context interaction layer. The similarity between sentence
      $tp_i$ and scene $z_i$ is computed by the TP--scene interaction
      layer.}
        \label{fig:model_subtask2}%
\end{figure}

\subsection{Identifying Turning Points in Screenplays}
\label{methodology:subtask2}

Identifying TPs in synopses serves as a testbed for validating some of
the assumptions put forward in this work, namely that turning points
mark narrative progression and can be identified automatically based
on their lexical makeup. Nevertheless, we are mainly interested in the
real-world scenario where TPs are detected in longer documents such as
screenplays. Screenplays are naturally segmented into scenes, which
often describe a self-contained event that takes place in one
location, and revolves around a few characters. We therefore assume
that scenes are suitable textual segments for signaling TPs in
screenplays. 

Unfortunately, we do not have any goldstandard information about TPs
in screenplays. We provide distant supervision by constructing noisy
labels based on goldstandard TP annotations in synopses (see the
description below). Given sentences labeled as TPs in a synopsis, we
identify scenes in the corresponding screenplay which are semantically
similar to them. We formulate this task as a binary classification
problem, where a sentence-scene pair is deemed either ``relevant'' or
``irrelevant'' for a given TP.

\paragraph{Distant Supervision}
\label{sec:data_construction}

Based on the screenwriting scheme of \newcite{Hauge:2017}, TPs are
expected to occur in specific parts of a screenplay (e.g., the Climax
is likely to occur towards the end). We exploit this knowledge as a
form of distant supervision.
We estimate the mean position for each TP using the gold standard
annotation of the plot synopses in our training set (normalized by the
synopsis length). The results are shown in
Table~\ref{tab:across_movies}, together with the TP positions
postulated by screenwriting theory. We observe that our estimates
agree well with the theoretical predictions, but also that some TPs
(e.g.,~TP2 and TP3) are more variable in their position than others
(e.g.,~TP1 and TP5). This leads us to the following hypothesis:
each TP is situated within a specific window in a screenplay. Scenes
that lie within the window are semantically related to the TP, whereas
all other scenes are unrelated. In experiments we calculate a window
$\mu \pm \sigma$ based on our data (see
Table~\ref{tab:across_movies}).

\begin{table}[t]
\small
\centering
\begin{tabular}{lccccc}
\thickhline
\textbf{} & {TP1} & {TP2} & {TP3} & {TP4} & {TP5} \\ \thickhline 
theory & 10.00 & 25.00 & 50.00 & 75.00 & \begin{tabular}[c]{@{}l@{}}94.50\end{tabular} \\  $\mu$ & 11.39 & 31.86 & 50.65 & 74.15 & 89.43 \\ 
 $\sigma$   & 6.72 & 11.26 & 12.15 & 8.40 & 4.74 \\ \thickhline
\end{tabular}
\caption{Expected TP position based on screenwriting
  theory; mean position $\mu$ and standard deviation~$\sigma$ in 
  goldstandard synopses of our training set.}
\label{tab:across_movies}
\end{table}

We compute scene representations based on the sequence of sentences
that comprise it using a BiLSTM equipped with an attention mechanism
(see Section~\ref{methodology:subtask1}). The final scene
representation~$s$ is the weighted sum of the representations of the
scene sentences. Next, the TP--scene interaction layer enriches scene representations with similarity values with each marked TP synopsis 
sentence $tp$ as shown in Equations~(\ref{interaction2})--(\ref{interaction4}).

We again augment the above-described base model with contextualized
sentence and scene representations using a synopsis and screenplay
encoder. The synopsis encoder is the same one used for our sentence-level
TP prediction task (see Section~\ref{single_plot_encoder}). The
screenplay encoder works in a similar fashion over scene
representations.

\paragraph{Topic-Aware Model (TAM)} TAM enhances our screenplay
encoder with information about topic boundaries.  Specifically, we
compute the representations of the left $lc_i$ and right $rc_i$
context window of the $i^{th}$ scene in the screenplay as described in
Section~\ref{sec:context_layer}. Next, we compute the final
representation $z_i$ of scene $sc_i$ by concatenating the
representations of the context windows $lc_i$ and $rc_i$ and the
current scene $sc_i$: $z_i = [lc_i ; sc_i ; rc_i]$.  There is no need
to compute the similarity between scenes and context windows here as
we now have goldstandard TP representations in the synopsis and employ
the TP--scene interaction layer for the computation of the similarity
between TPs and enriched scene representations $z_i$. Hence, we
directly calculate in this layer a scene-level feature vector that
encodes information about the scene, its similarity to TP sentences,
and whether these function as boundaries between topics in the
screenplay.

\paragraph{Entity-Specific information} We can also employ an
  entity-specific encoder (see Section~\ref{sec:es_encoder}) for the
  representing the synopsis and scene sentences. Again, generic and
  entity-specific representations are combined via concatenation.

\subsection{End-to-end TP Identification}
\label{methodology:subtask3}

Our ultimate goal is to identify TPs in screenplays without assuming
any goldstandard information about their position in the synopsis. We
address this with an end-to-end model which first predicts the
sentences that act as TPs in the synopsis (e.g.,~TAM in
Section~\ref{methodology:subtask1}) and then feeds these predictions
to a model which identifies the corresponding TP scenes (e.g.,~TAM in
Section~\ref{methodology:subtask2}).

\section{Experimental Setup} 
\label{sec:setup}

\paragraph{Training} We used the Universal Sentence Encoder (USE;
\citealt{cer2018universal}) as a pre-trained sentence encoder for all
models and tasks; its performance was superior to~BERT
\cite{devlin2018bert} and other related pre-trained encoders (for more
details, see the Appendix). Since the binary labels in both prediction
tasks are imbalanced, we apply class weights to the loss function of
our models. We weight each class by its inverse frequency in the
training set (for more implementation details, see the Appendix).

\paragraph{Inference}
During inference in our first task (i.e.,~identification of TPs in
synopses), we select one sentence per TP. Specifically, we want to
track the five sentences with the highest posterior probability of
being TPs and sequentially assign them TP labels based on their
position. However, it is possible to have a cluster of neighboring
sentences with high probability, even though they all belong to the
same TP. We therefore constrain the sentence selection for each TP
within the window of its expected position, as calculated in the
distribution baseline (Section~\ref{sec:data_construction}).

For models which predict TPs in screenplays, we obtain a probability
distribution over all scenes in a screenplay indicating how relevant
each is to the TPs of the corresponding plot synopsis. We find the
peak of each distribution and select a neighborhood of scenes around
this peak as TP-relevant ones. Based on the goldstandard annotation,
each TP corresponds to 1.77 relevant scenes on average
(StDev~1.23). We therefore consider a neighborhood of three relevant
scenes per TP.

\begin{table}[t]
  \centering
   \begin{subtable}{0.45\textwidth}
        \small
       \centering
      \begin{tabular}{lcc}
        \thickhline
       & \multicolumn{1}{c}{\begin{tabular}[c]{@{}c@{}}{$TA$}\end{tabular}} & \multicolumn{1}{c}{\begin{tabular}[c]{@{}c@{}}{$D$}\end{tabular}} \\ \thickhline
         \begin{tabular}[c]{@{}l@{}} Baseline \end{tabular} &
                                                                   31.00 & 9.65 (4.41) \\
         \begin{tabular}[c]{@{}l@{}} CAM \end{tabular} & 33.00 & 7.44 (8.09) \\

        \begin{tabular}[c]{@{}l@{}} TAM \end{tabular}  & 36.00 & 7.11 (7.98) \\
        \begin{tabular}[c]{@{}l@{}} \hspace{1ex}+ TP views \end{tabular}  & \textbf{39.00} & \textbf{6.52} (\textbf{7.72}) \\
        \begin{tabular}[c]{@{}l@{}} \hspace{1ex}+ entities \end{tabular}  & 38.00 & 6.91 (7.65) \\ \thickhline
    \end{tabular}
    \vspace{-0.4em}
    \caption{Development set}
    \label{tab:plots_results}
   \end{subtable}%
    \vspace{1em}
    \begin{subtable}{0.45\textwidth}
        \small
       \centering
        \begin{tabular}{lcc}
        \thickhline
& \multicolumn{1}{c}{\begin{tabular}[c]{@{}c@{}}{$TA$}\end{tabular}}
  & \multicolumn{1}{c}{\begin{tabular}[c]{@{}c@{}}{$D$} \end{tabular}} \\ \thickhline
          Random & 2.00 & 37.79 (25.33) \\
          Theory baseline & 22.00 & 7.47 (6.75) \\
          Distribution baseline & 28.00 & {7.28} ({6.23}) \\
           \begin{tabular}[c]{@{}l@{}} TAM \end{tabular} & 34.67 &
           \textbf{6.80} (\textbf{5.19})\\
        \begin{tabular}[c]{@{}l@{}} \hspace{1ex}+ TP views \end{tabular} & {38.57} & 7.47 (7.48) \\
        \begin{tabular}[c]{@{}l@{}} \hspace{1ex}+ entities \end{tabular}  & \textbf{41.33} & {7.30} ({7.21})  \\ \thickhline
        \begin{tabular}[c]{@{}l@{}}\textit{Human agreement}\end{tabular} & {64.00}  & {4.30} ({3.43})  \\ \thickhline
        \end{tabular}
 \vspace{-0.4em}
\caption{Test set}
	\label{tab:plots_results_test}
   \end{subtable}
       \caption{Identification of TPs in
         plot synopses; results are shown in percent ($TA$: mean
         Total Agreement;
         $D$:~annotation distance; standard deviation in brackets).}
\end{table}

\begin{table}[t]
\centering
\small
\begin{tabular}{@{}llllll@{}}
\thickhline
\multicolumn{1}{c}{TAM} & {TP1}  & {TP2}   & {TP3}   & {TP4}   & {TP5}  \\ \thickhline
+ TP views  & \textbf{6.09} & 9.45 & 10.72 & 6.91 & 4.26 \\
+ entities  & 7.18 & \textbf{9.35} & \textbf{9.86} & \textbf{5.23} & \textbf{3.48} \\ \thickhline
\textit{Human agreement} & {3.33} & {5.00}  & {10.58} & {1.07}  & {1.53} \\ \thickhline
\end{tabular}
\caption{Mean annotation distance~$D$  (test set); results are
  shown per TP on the synopsis identification task.}
\label{tab:plot_results_per_tp}
\end{table}

\section{Results} 
\label{sec:results}

\textbf{TP Identification in Synopses}
Table~\ref{tab:plots_results} reports our results on the development
set (we extracted 20~movies from the original training set) which aim
at comparing various model instantiations for the TP identification
task. Specifically, we report the performance of a baseline model
which is neither context-aware nor utilizes topic boundary information
against CAM and TAM. We also show two variants of TAM enhanced with
TP-specific encoders (+ TP views) and entity-specific information (+
entities). Model performance is measured using the evaluation metrics
of Total Agreement ($TA$) and annotation distance ($D$), normalized by
synopsis length (equation~(\ref{eq:distance_1})).

The baseline model presents the lowest performance among all variants
which suggests that state-of-the-art sentence representations on their
own are not suitable for our task. Indeed, when contextualizing the
synopsis sentences via a BiLSTM layer we observe an absolute increase
of 4.00\% in terms of~$TA$. Moreover, the addition of a context
interaction layer (see TAM row in Table~\ref{tab:plots_results}) yields
an absolute $TA$~improvement of 4.00\% compared to CAM.  Combining
different TP views further improves by 3.00\%, reaching a $TA$
of~39.00\%, and reducing $D$ to~6.52\%.

\begin{figure}[t]
    \centering
    \includegraphics[width=\columnwidth]{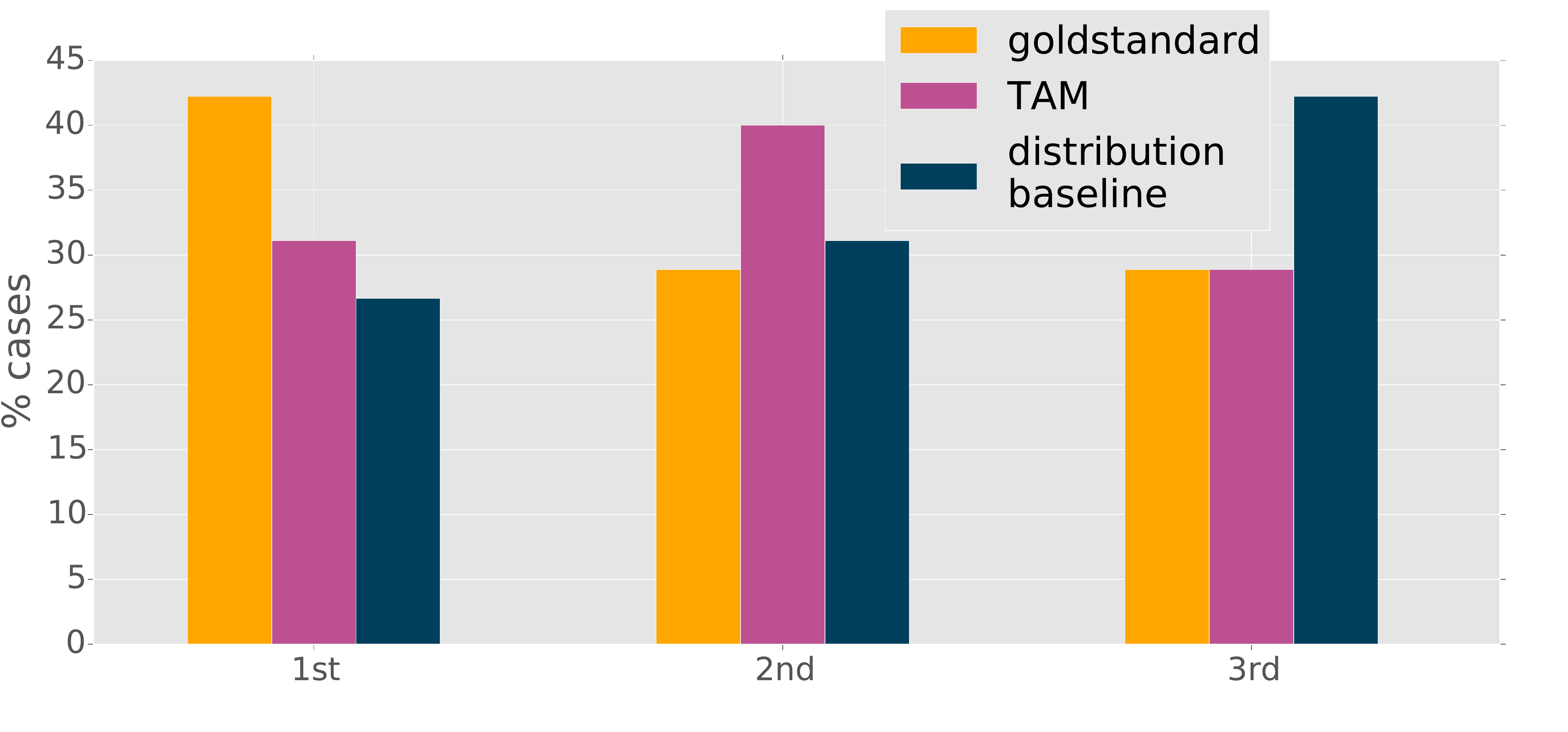}
    \caption{Rankings (shown as proportions) of synopsis highlights
      produced by aggregating goldstandard TP annotations, those
      predicted by the distribution baseline, and our model (TAM + TP
      views).}
    \label{fig:model_amt}
\end{figure}

\begin{figure*}[t]
\tiny
    \centering
\begin{tabular}{c@{}c@{}c}
\includegraphics[width=0.33\textwidth]{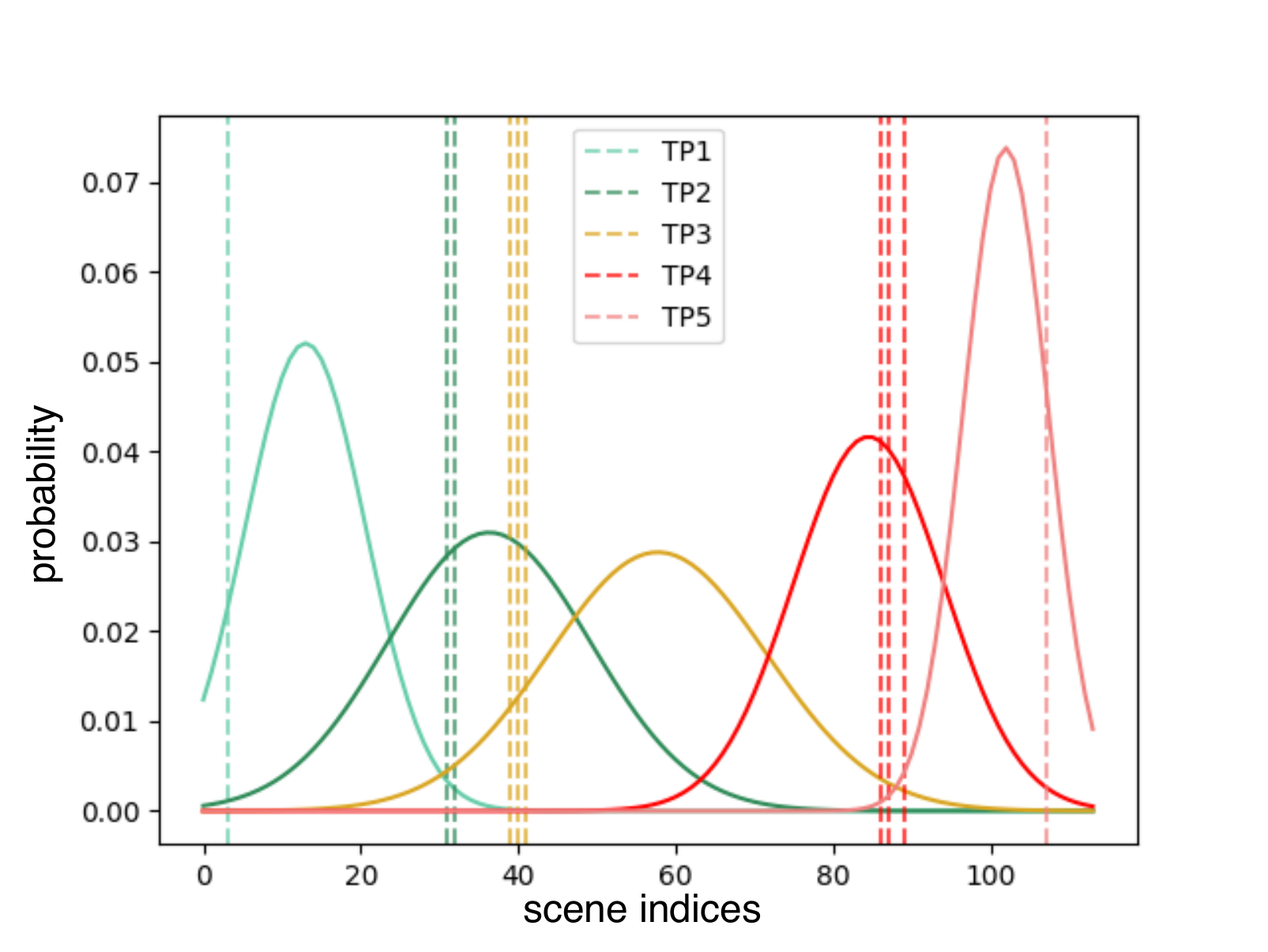}
& \includegraphics[width=0.33\textwidth]{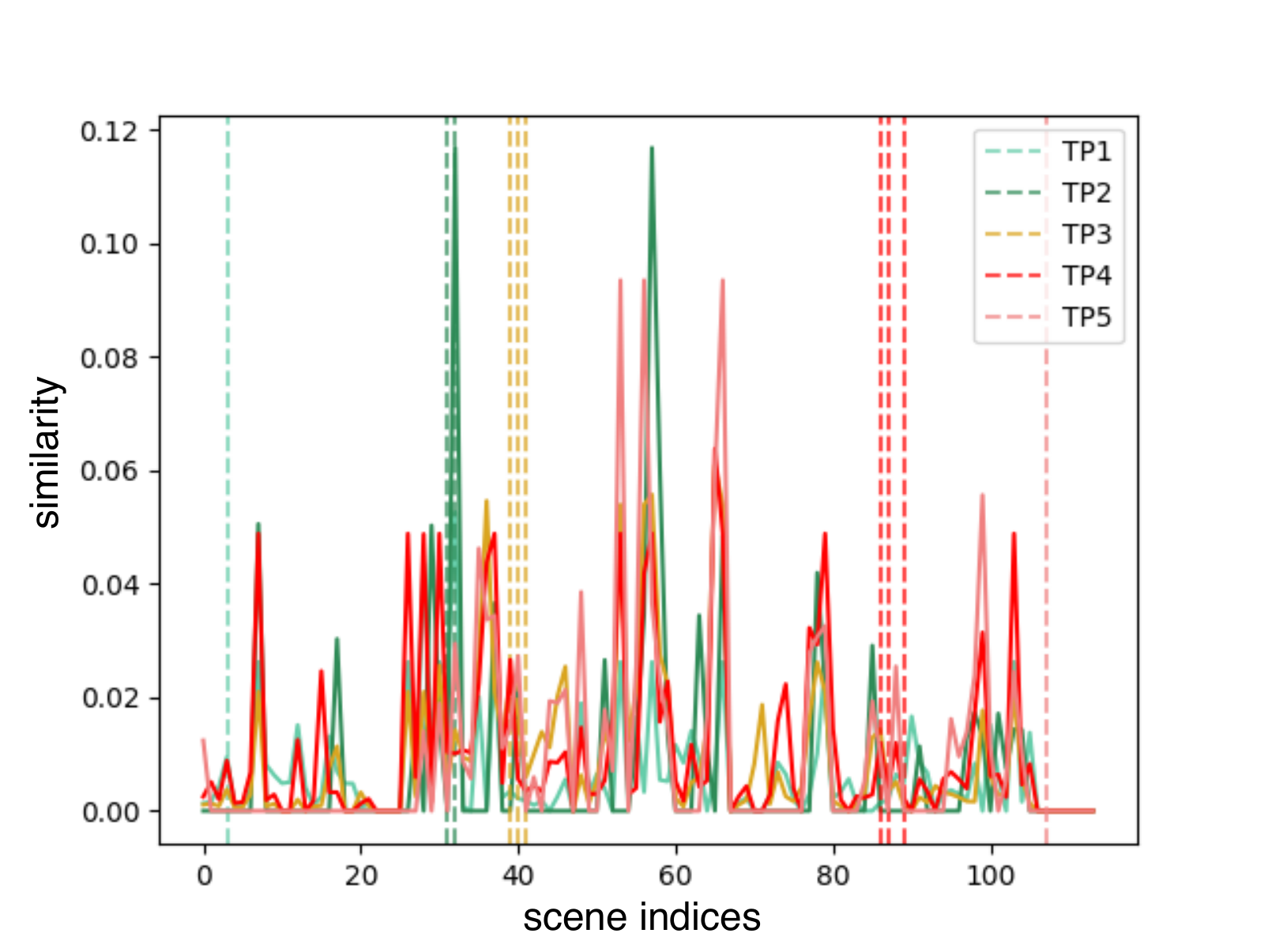}
&  \includegraphics[width=0.33\textwidth]{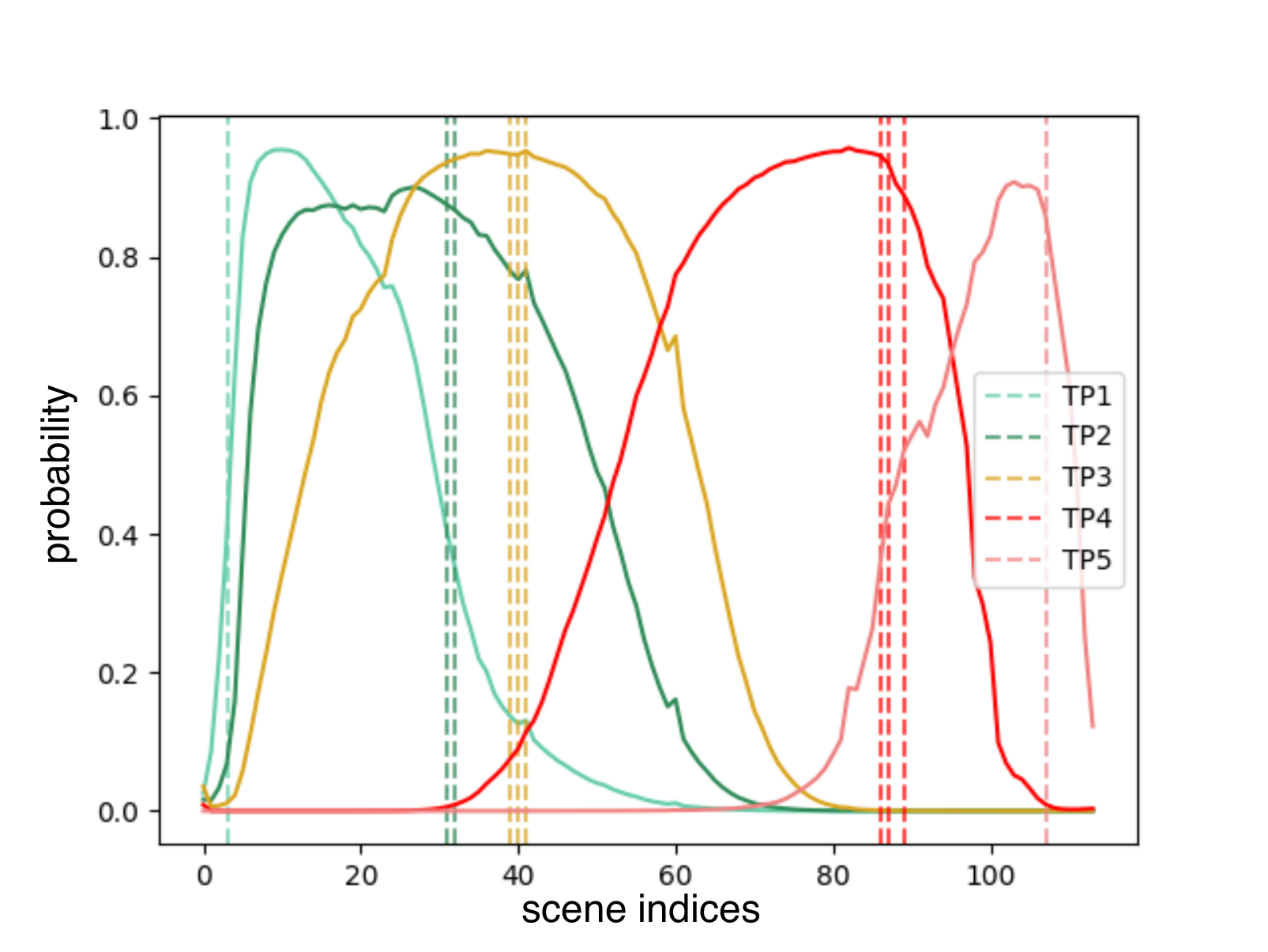}\\
 {{\small Distribution baseline}} \label{fig:example_baseline} &
 {{\small tf*idf similarity}} \label{fig:example_baseline} & {{\small TAM}}    
        \label{fig:example_mcam} \\
\end{tabular}
\caption
{\small Probability distributions over the scenes of the screenplay
  for the movie ``Juno''; x-axis: scene indices, y-axis: probability
  that the scene is relevant to a specific TP. Vertical dashed lines
  are goldstandard TP scenes.}
    \label{fig:subtask_2_example}
\end{figure*}

Table~\ref{tab:plots_results_test} shows our results on the test set.
We compare TAM, our best performing model against two strong
baselines. The first one selects sentences that lie on the expected
positions of TPs according to screenwriting theory; while the second
one selects sentences that lie on the peaks of the empirical TP
distributions in the training set
(Section~\ref{sec:data_construction}). As we can see, TAM (+ TP views)
achieves a $TA$ of~38.57\% compared to 22.00\% for the distribution
baseline. And although entity-specific information does not have much
impact on the development set, it yields a 2.76\% improvement on the
test set. A detailed break down of results per TP is given in
Table~\ref{tab:plot_results_per_tp}. Interestingly, our model
resembles human behavior (see row \emph{Human agreement}): TPs 1, 4,
and 5 are easiest to distinguish, whereas TPs 2 and 3 are hardest and
frequently placed at different points in the
synopsis.

We also conducted a human evaluation experiment on Amazon Mechanical
Turk (AMT). AMT workers were presented with a synopsis and
``highlights'', i.e., five sentences corresponding to TPs. We obtained
highlights from goldstandard annotations, the distribution baseline,
and TAM (+ TP views). AMT workers were asked to read the synopsis and
rank the highlights from best to worst according to the following
criteria: (1) the quality of the plotline that they form; (2) whether
they include the most important events and plot twists of the movie;
and (3) whether they provide some description of the events in the
beginning and end of the movie.  In Figure~\ref{fig:model_amt} we
show, proportionally, how often our participants ranked each model
1st, 2nd, and so on. Perhaps unsurprisingly, goldstandard TPs were
considered best (and ranked 1st 42\% of the time). TAM is ranked best
30\% of the time, followed by the distribution baseline which was only
ranked first 26\% of the time.  Overall, the average ranking positions
for the goldstandard, TAM, and the baseline are 1.87, 1.98, and 2.16,
respectively. Human evaluation therefore validates that our model
outperforms the position-based baselines.

\begin{table}[t]
\centering
\small
\begin{tabular}{lccc}
\thickhline
 & \begin{tabular}[c]{@{}c@{}} {$TA$} \end{tabular} & \begin{tabular}[c]{@{}c@{}} {$PA$} \end{tabular} & \begin{tabular}[c]{@{}c@{}} {$D$} \end{tabular} \\ \thickhline
 Theory baseline & 8.66 & 10.67  & 10.45 (9.14)     \\ 
Distribution baseline & 6.67 & 9.33  & 10.84 (8.94)     \\ 
tf*idf similarity &  0.74 & 1.33 & 53.07 (31.83)   \\ 
tf*idf + distribution & 4.44 & 6.67 & 13.33 (11.51)  \\ 
\begin{tabular}[c]{@{}l@{}} CAM \end{tabular} & 11.11  & 16.00   & 10.23 (11.23)  \\ 
\begin{tabular}[c]{@{}l@{}} \hspace*{1ex}+ entities \end{tabular} & \textbf{14.18} & \textbf{17.33} & 12.77 (12.61)  \\ 
\begin{tabular}[c]{@{}l@{}} TAM \end{tabular} & 10.63 & 13.33  &  \textbf{8.94} (\textbf{9.39})    \\ 
\begin{tabular}[c]{@{}l@{}} \hspace*{1ex}+ entities  \end{tabular} & 10.63 & 13.33 & 10.15 (10.56)    \\ 
\begin{tabular}[c]{@{}l@{}} TAM End2end \end{tabular} & 7.87 & 9.33 & 10.16 (10.74) \\ \thickhline
\begin{tabular}[c]{@{}l@{}} \textit{Human agreement} \end{tabular}& \textit{35.48} & \textit{56.67} & \textit{1.48} (\textit{2.93})  \\ \thickhline
\end{tabular}
\caption{Identification of TPs in screenplays; results are shown in
  percent using five-fold cross validation ($TA$:  mean Total
  Agreement; $PA$: Partial Agreement; $D$:  annotation
  distance $D$; standard deviation  in brackets).}
\label{tab:script_results_test}
\end{table}

\paragraph{TP Identification in Screenplays} Our results are
summarized in Table~\ref{tab:script_results_test}. For this task, we
performed five-fold crossvalidation over our original goldstandard set
to obtain a test-development split (recall we do not have goldstandard
annotations for training). We report Total Agreement ($TA$), Partial
Agreement ($PA$), and annotation distance $D$, normalized by
screenplay length
(Equations~(\ref{eq:total_agreement})--(\ref{eq:distance_script_1})).

Aside from the theory and distribution-based baselines, we also
experimented\footnote{Common segmentation approaches such as
  TextTiling \cite{hearst1997texttiling} perform poorly on our task
  and we do not report them due to space constraints.} with a
common IR baseline which considers TP synopsis sentences as queries
and retrieves a neighborhood of semantically similar scenes from the
screenplay using tf*idf similarity.  Specifically, we compute the
maximum tf*idf similarity for all sentences included in the respective
scene. We empirically observed that tf*idf's behavior can be erratic
selecting scenes in completely different sections of the screenplay,
and therefore constrain it by selecting scenes only within the windows
determined by the position distributions ($\mu \pm \sigma$) for each
TP. As far as our own models are concerned, we report results with
goldstandard TP labels for CAM and TAM on their own and enriched with
entity information. We also built and end-to-end system based on TP
predictions from TAM.

As can be seen in Table~\ref{tab:script_results_test}, tf*idf
approaches perform worse than position-related baselines. Overall,
similar vocabulary across scenes and mentions of the same entities
throughout the screenplay make tf*idf approaches insufficient for our
tasks. The best performing model is TAM confirming our hypothesis that
TPs are not just isolated key events, but also mark boundaries between
thematic units and, therefore, segmentation-inspired approaches can be
beneficial for the task. Results for entities are somewhat mixed; for
CAM, the entity-specific information improves $TA$ and $PA$ but
increases $D$, while it does not seem to make much difference for TAM.
The performance of the end-to-end TAM model drops slightly compared to
the same model using goldstandard TP annotations. However, it still
remains competitive against the baselines, indicating that tracking
TPs in screenplays fully automatically is feasible.

In Figure~\ref{fig:subtask_2_example}, we visualize the posterior
distribution of various models over the scenes of the screenplay for
the movie ``Juno''. The first panel shows the distribution baseline
alongside goldstandard TP scenes (vertical lines). We observe that the
distribution baseline provides a good approximation of relevant TP
positions (which validates its use in the construction of noisy
labels, Section~\ref{sec:data_construction}), even though it is not
always accurate. For example, TPs 1 and 3 lie outside the expected
window in ``Juno''.

The second panel presents the TP predictions according to tf*idf
similarity. We observe that scenes located in entirely different parts
of the screenplay present high similarity scores with respect to a
given TP due to vocabulary uniformity and mentions of the same
entities throughout the screenplay.
In the next panel we present the predictions of TAM. Adding synopsis
and screenplay encoders yields smoother distributions increasing the
probability of selecting TP scenes inside distinct regions of the
screenplay, with sharper peaks and higher
confidence.

\section{Conclusions} 
\label{sec:conclusions}

We proposed the task of turning point identification in screenplays as
a means of analyzing their narrative structure.  We demonstrated that
automatically identifying a sequence of key events and segmenting the
screenplay into thematic units is feasible via an end-to-end neural
network model.  In future work, we will investigate the usefulness of
TPs for summarization and question answering. We will also scale the
TRIPOD dataset and move to a multi-modal setting where TPs are
identified directly in video data.

\section*{Acknowledgments}
We thank the anonymous reviewers for their feedback.  We gratefully
acknowledge the support of the European Research Council (Lapata;
award 681760, ``Translating Multiple Modalities into Text'') and of
the Leverhulme Trust (Keller; award IAF-2017-019).

\bibliography{references}
\bibliographystyle{acl_natbib}

\appendix

\section{Model Details}

\paragraph{Synopsis Encoder} In all tasks we use a synopsis encoder in
order to contextualize the sentences in the synopsis. We employ an
LSTM network as the synopsis encoder which produces sentence
representations $h_1,h_2,\dots,h_T$, where $h_i $ is the hidden state
at time-step $i$, summarizing all the information of the synopsis up
to the $i$-th sentence.  We use a Bidirectional LSTM (BiLSTM) in order
to get sentence representations that summarize the information from
both directions. A BiLSTM consists of a forward LSTM $
\overrightarrow{f} $ that reads the synopsis from $p_1$ to $p_N$ and a
backward LSTM $ \overleftarrow{f} $ that reads it from $p_N$ to
$p_1$. We obtain the final representation $cp_i$ for a given synopsis
sentence $p_i$ by concatenating the representations from both
directions, $ cp_i = h_i = [\overrightarrow{h_i} ;
\overleftarrow{h_i}], \quad h_i \in R^{2S}, $ where $S$ denotes the
size of each LSTM.

\paragraph{Entity-Specific Encoder} This encoder is used to evaluate
the contribution of entity-specific information to the performance
of our models. We use a word embedding layer to project words
$w_1,w_2,\dots,w_T$ of the $i^{th}$ synopsis sentence~$p_i$ to a
continuous vector space $R^E$, where $E$ the size of the embedding
layer. This layer is initialized with pre-trained entity
embeddings. Next, we use a BiLSTM as described in the case of the
synopsis encoder.  On top of the LSTM, we add an attention mechanism,
which assigns a weight $a_i$ to each word representation $h_i$. We
compute the entity-specific representation $pe_i$ of the $i^{th}$ plot
sentence as the weighted sum of word representations:
\begin{gather}
e_j = \tanh(W_h h_j + b_h)\label{eq:att_ei_ent}, \quad e_j \in [-1,1]
\end{gather}
\begin{gather}
    a_j = \dfrac{\exp(e_j)}{\sum_{t=1}^{T} \exp(e_t)}\label{eq:att_ai_ent}, \quad \sum_{j=1}^{T} a_j = 1 \\
    pe_i = \sum_{j=1}^{T} a_jh_j \label{eq:att_r_ent}, \quad e \in R^{2S}
\end{gather}
where $W_h$ and~$b_h$ are the attention layer's weights.

\begin{table*}[t]
\centering
\small
\begin{tabular}{@{}p{16cm}@{}}
\thickhline
\multicolumn{1}{c}{Goldstandard} \\\thickhline
   \tabitem Sixteen-year-old Minnesota high-schooler Juno MacGuff discovers she is pregnant with a child fathered by her friend and longtime admirer, Paulie Bleeker. \\
 \tabitem All of this decides her against abortion, and she decides to give the baby up for adoption. \\
  \tabitem With Mac, Juno meets the couple, Mark and Vanessa Loring (Jason Bateman and Jennifer Garner), in their expensive home and agrees to a closed adoption. \\
  \tabitem Juno watches the Loring marriage fall apart, then drives away and breaks down in tears by the side of the road. \\
  \tabitem Vanessa comes to the hospital where she joyfully claims the newborn boy as a single adoptive mother.
 \\ \thickhline
\multicolumn{1}{c}{TAM (+ TP views)} \\ \thickhline
  \tabitem Going to a local clinic run by a women's group, she encounters outside a school mate who is holding a rather pathetic one-person Pro-Life vigil. \\
 \tabitem With Mac, Juno meets the couple, Mark and Vanessa Loring (Jason Bateman and Jennifer Garner), in their expensive home and agrees to a closed adoption. \\
 \tabitem Juno and Leah happen to see Vanessa in a shopping mall being completely at ease with a child, and Juno encourages Vanessa to talk to her baby in the womb, where it obligingly kicks for her. \\
 \tabitem Juno watches the Loring marriage fall apart, then drives away and breaks down in tears by the side of the road. \\
 \tabitem The film ends in the summertime with Juno and Paulie playing guitar and singing together, followed by a kiss.\\ \thickhline
\multicolumn{1}{c}{Distribution baseline} \\ \thickhline
\tabitem Once inside, however, Juno is alienated by the clinic staff's authoritarian and bureaucratic attitudes. \\
 \tabitem Juno visits Mark a few times, with whom she shares tastes in punk rock and horror films. \\
 \tabitem Not long before her baby is due, Juno is again visiting Mark when their interaction becomes emotional. \\
 \tabitem Juno then tells Paulie she loves him, and Paulie's actions make it clear her feelings are very much reciprocated. \\
 \tabitem Vanessa comes to the hospital where she joyfully claims the newborn boy as a single adoptive mother. \\ \thickhline
\end{tabular}
\caption{Highlights for the movie ''Juno'':  goldstandard annotations and
  predicted TPs for TAM (+ TP views) and distribution baseline.}
\label{tab:example_juno}
\end{table*}

\section{Implementation Details}

\paragraph{Pre-trained Sentence Encoder} The performance of our models
depends on the initial sentence representations. We
experimented with using the large BERT model \cite{devlin2018bert} and
the Universal Sentence Encoder (USE) \cite{cer2018universal} as
pre-trained sentence encoders in all tasks. Intuitively, we expect USE
to be more suitable, since it was trained in textual similarity tasks
which are more relevant to ours. Experiments on the development set
confirmed our intuition.  Specifically, on the screenplay TP
prediction task, annotation distance~$D$ dropped from~17.00\% to
10.04\% when employing USE instead of the BERT embeddings in the CAM
version of our architecture.

\paragraph{Hyper-parameters} 
We used the Adam algorithm \cite{kingma2014adam} for optimizing our
networks. After experimentation, we chose an LSTM with 32 neurons (64
for the BiLSTM) for the synopsis encoder in the first task and one
with 64 neurons for the encoder in the second task. For the context
interaction layer, the window~$l$ was set to two sentences for the
first task and 20\% of the screenplay length for the second task.  For
the entity encoder, an embedding layer of size 300 was initialized
with the Wikipedia2Vec pre-trained word embeddings
\cite{yamada2018wikipedia2vec} and remained frozen during
training. The LSTM of the encoder had 32 and 64 neurons for the first
and second tasks, respectively. Finally, we also added a dropout
of~0.2. For developing our models we used PyTorch
\cite{paszke2017automatic}.

\paragraph{Data Augmentation} We used multiple annotations for
training for movies where these were available and considered
reliable.  The reasons for this are twofold. Firstly, this allowed us
to take into account the subjective nature of the task during
training; and secondly, it increased the size of our dataset, which
contains a limited number of movies. Specifically, we added triplicate
annotations for 17 movies and duplicate annotations for 5 movies.

\begin{table*}[t]
\centering
\small
\begin{tabular}{@{}p{15cm}@{}}
\thickhline
\multicolumn{1}{c}{Goldstandard} \\ \thickhline
   \tabitem On the night the two move into the home, it is broken into by Junior, the previous owner's grandson; Burnham, an employee of the residence's security company; and Raoul, a ski mask-wearing gunman recruited by Junior. \\
     \tabitem Before the three can reach them, Meg and Sarah run into the panic room and close the door behind them, only to find that the burglars have disabled the telephone. \\
     \tabitem To make matters worse, Sarah, who has diabetes, suffers a seizure. \\
     \tabitem Sensing the potential danger to her daughter, Meg lies to the officers and they leave. \\
     \tabitem After a badly injured Stephen shoots at Raoul and misses, Raoul disables him and prepares to kill Meg with the sledgehammer, but Burnham, upon hearing Sarah's screams of pain, returns to the house and shoots Raoul dead, stating, "You'll be okay now", to Meg and her daughter before leaving.
 \\ \thickhline
\multicolumn{1}{c}{TAM (+ TP views)} \\ \thickhline
  \tabitem On the night the two move into the home, it is broken into by Junior, the previous owner's grandson; Burnham, an employee of the residence's security company; and Raoul, a ski mask-wearing gunman recruited by Junior. \\
 \tabitem Before the three can reach them, Meg and Sarah run into the panic room and close the door behind them, only to find that the burglars have disabled the telephone.\\
 \tabitem Her emergency glucagon syringe is in a refrigerator outside the panic room. \\
 \tabitem As Meg throws the syringe into the panic room, Burnham frantically locks himself, Raoul, and Sarah inside, crushing Raoul's hand in the sliding steel door. \\
 \tabitem After a badly injured Stephen shoots at Raoul and misses, Raoul disables him and prepares to kill Meg with the sledgehammer, but Burnham, upon hearing Sarah's screams of pain, returns to the house and shoots Raoul dead, stating, "You'll be okay now", to Meg and her daughter before leaving.
\\ \thickhline
\multicolumn{1}{c}{Distribution baseline} \\ \thickhline 
\tabitem On the night the two move into the home, it is broken into by Junior, the previous owner's grandson; Burnham, an employee of the residence's security company; and Raoul, a ski mask-wearing gunman recruited by Junior. \\
 \tabitem Unable to seal the vents, Meg ignites the gas while she and Sarah cover themselves with fireproof blankets, causing an explosion which vents into the room outside and causes a fire, injuring Junior. \\
 \tabitem To make matters worse, Sarah, who has diabetes, suffers a seizure. \\
 \tabitem While doing so, he tells Sarah he did not want this, and the only reason he agreed to participate was to give his own child a better life. \\
 \tabitem As the robbers attempt to leave, using Sarah as a hostage, Meg hits Raoul with a sledgehammer and Burnham flees. \\ \thickhline
\end{tabular}
\caption{Highlights for the movie ''Panic Room'': goldstandard annotations
  and the predicted TPs for TAM (+ TP views) and distribution baseline.}
\label{tab:example_panic_room}
\end{table*}

\begin{table*}[h!]
\centering
\small
\begin{tabular}{@{}p{15cm}@{}}
\thickhline
\multicolumn{1}{c}{Goldstandard} \\ \thickhline
   \tabitem Manager Stuart Ullman warns him that a previous caretaker developed cabin fever and killed his family and himself. \\
 \tabitem Hallorann tells Danny that the hotel itself has a "shine" to it along with many memories, not all of which are good. \\
 \tabitem After she awakens him, he says he dreamed that he had killed her and Danny. \\
 \tabitem Jack begins to chop through the door leading to his family's living quarters with a fire axe. \\
 \tabitem Wendy and Danny escape in Hallorann's snowcat, while Jack freezes to death in the hedge maze. \\ \thickhline
\multicolumn{1}{c}{TAM (+TP views)} \\ \thickhline
  \tabitem Jack's wife, Wendy, tells a visiting doctor that Danny has an imaginary friend named Tony, and that Jack has given up drinking because he had hurt Danny's arm following a binge. \\
 \tabitem Hallorann tells Danny that the hotel itself has a "shine" to it along with many memories, not all of which are good. \\
 \tabitem Danny starts calling out "redrum" frantically and goes into a trance, now referring to himself as "Tony". \\
 \tabitem When Wendy sees this in the bedroom mirror, the letters spell out "MURDER". \\
 \tabitem Wendy and Danny escape in Hallorann's snowcat, while Jack freezes to death in the hedge maze.
\\ \thickhline
\multicolumn{1}{c}{Distribution baseline} \\ \thickhline
\tabitem Jack's wife, Wendy, tells a visiting doctor that Danny has an imaginary friend named Tony, and that Jack has given up drinking because he had hurt Danny's arm following a binge. \\
 \tabitem Jack, increasingly frustrated, starts acting strangely and becomes prone to violent outbursts. \\
 \tabitem Jack investigates Room 237, where he encounters the ghost of a dead woman, but tells Wendy he saw nothing. \\
 \tabitem When Wendy sees this in the bedroom mirror, the letters spell out "MURDER". \\
 \tabitem He kills Hallorann in the lobby and pursues Danny into the hedge maze. \\ \thickhline
\end{tabular}
\caption{Highlights for the movie ''The Shining'':  goldstandard
  annotations and the predicted TPs TAM (+ TP views) and distribution baseline.}
\label{tab:example_shining}
\end{table*}

\section{Example Output: TP Identification in Synopses}

As mentioned in Section 6, we also conducted a human
evaluation experiment, where highlights were extracted by combining
the five sentences labeled as TPs the synopsis. In Tables
\ref{tab:example_juno}, \ref{tab:example_panic_room}, and
\ref{tab:example_shining}, we present the highlights presented to the
AMT workers for the movies ''Juno'', ''Panic Room'', and ''The
Shining'', respectively. For each movie we show the goldstandard
annotations alongside with the predicted TPs for TAM (+ TP views) and
the distribution baseline, which is the strongest performing baseline
with respect to the automatic evaluation results.

Overall, we observe that goldstandard highlights describe the plotline
of the movie, contain a first introductory sentence,
some major and intense events, and a last sentence that
describes the ending of the story.

The distribution baseline is able to
predict a few goldstandard TPs by only considering the relative
position of the sentences in the synopsis. This observation validates
the screenwriting theory: TPs, or more generally  important events
that determine the progression of the plot, are consistently
distributed in specific parts of a movie. However, when
the distribution baseline cannot predict the exact TP
sentence, it might select one that describes  irrelevant events of
minor importance (e.g.,~TP4 for ''Panic Room'' is a detail about a
secondary character instead of a major setback and highly intense
event in the movie). 

Finally, our own model seems to be able to predict some goldstandard
TP sentences, as demonstrated during the automatic
evaluation. However, we also observe here that even when it does not
select the goldstandard TPs, the predicted ones describe important
events in the movie that have some desired characteristics. In
particular, for the movie ''Juno'' the climax (TP5) is the moment of
resolution, where Vanessa decides to adopt the baby after all the
setbacks and obstacles. Even though our model does not predict this
sentence, it does select one that reveals information about the ending
of the movie. An other such example is the movie ''Panic Room'', where
the point of no return (TP3) is not correctly predicted, but the
selected sentence refers to the same event.

\end{document}